\documentclass[runningheads,a4paper]{llncs}

\usepackage{amssymb}
\usepackage{graphicx}
\usepackage{epstopdf}
\usepackage{multirow}
\usepackage{algpseudocode}
\usepackage[ruled, vlined, linesnumbered, resetcount]{algorithm2e}
\usepackage{cite}
\usepackage{subfig}
\usepackage{url}

\urldef{\mailsa}\path|{tegjyotsingh.sethi, mehmedkantardzic}@louisville.edu |    
\newcommand{\keywords}[1]{\par\addvspace\baselineskip
\noindent\keywordname\enspace\ignorespaces#1}

\begin{document}

\mainmatter  

\title{\lq Security Theater\rq: On the Vulnerability of Classifiers to Exploratory Attacks}

\titlerunning{On the Vulnerability of Classifiers to Exploratory Attacks}

%
%
\author{ Tegjyot Singh Sethi\inst{1} \and Mehmed Kantardzic\inst{1} \and Joung Woo Ryu\inst{2}}

\authorrunning{T. Sethi et al.}

\institute{Data Mining Lab, University of Louisville, USA\\
\mailsa\\
\and
Onycom Inc., Seoul, Republic of Korea \\
\email ryu0914@gmail.com
}

\toctitle{Lecture Notes in Computer Science}
\tocauthor{Authors' Instructions} 
\maketitle

\begin{abstract}

The increasing scale and sophistication of cyber-attacks has led to the adoption of machine learning based classification techniques, at the core of cybersecurity systems. These techniques promise scale and accuracy, which traditional rule/signature based methods cannot. However, classifiers operating in adversarial domains are vulnerable to evasion attacks by an adversary, who is capable of learning the behavior of the system by employing intelligently crafted probes. Classification accuracy in such domains provides a false sense of security, as detection can easily be evaded by carefully perturbing the input samples. In this paper, a generic data driven framework is presented, to analyze the vulnerability of classification systems to black box probing based attacks. The framework uses an \textit{exploration-exploitation} based strategy, to understand an adversary's point of view of the attack-defense cycle. The adversary assumes a black box model of the defender's classifier and can launch indiscriminate attacks on it, without information of the defender's model type, training data or the domain of application. Experimental evaluation on 10 real world datasets demonstrates that even models having high perceived accuracy ($>$90\%), by a defender, can be effectively circumvented with a high evasion rate ($>$95\%, on average). The detailed attack algorithms, adversarial model and empirical evaluation, serve as a background for developing secure machine learning based systems. 

\end{abstract}

\section{Introduction}
\label{sec:intro}

The Big Data revolution has fueled the development of scalable and practical machine learning systems, which has in turn led to their widespread adaptation and popularity. The domain of cybersecurity has also recognized the need for a data driven solution \cite{abramson2015toward, akhtar2011robustness, barreno2006can, biggio2014security, zhou2016modeling} , owing to the increased scale and sophistication of attacks in recent times\footnote{\url{https://www.statista.com/chart/2540/data-breaches/}}. Although the use of machine learning techniques has found early success in many cybersecurity applications (such as spam filtering \cite{kantchelian2013approaches}, CAPTCHA systems \cite{d2014avatar}, and intrusion detection \cite{biggio2014security}), its own vulnerabilities have mostly been overlooked. Machine learning systems were designed under the assumption of stationarity, i.e. the training and the testing dataset should be identically and independently distributed \cite{ditzler2015learning}. This assumption is often violated in cybersecurity domains, as the systems operate in a dynamic and adversarial environment \cite{barreno2006can, abramson2015toward}. 

\begin{figure}[t]
\centering
\includegraphics[width=0.7\textwidth]{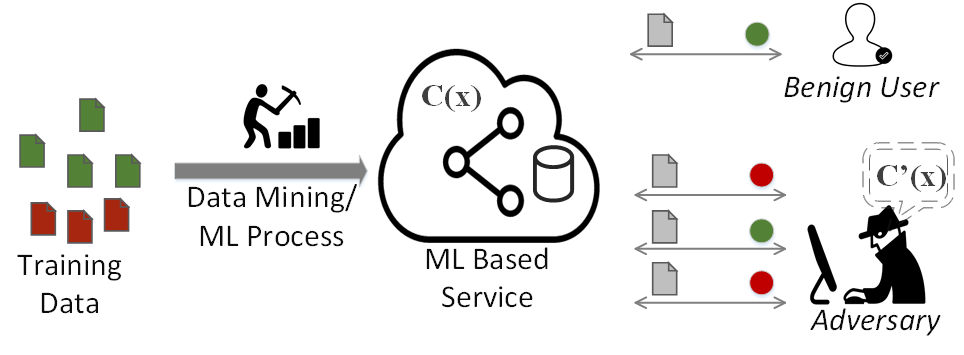}
\caption{Classifier systems in adversarial environments. An adversary making probes to the model $C(x)$ can use active learning to arrive at their own understanding of the model, as $C'(x)$. Probes are made through the same channels as by other users accessing the ML service, and as such can be indistinguishable.  }
\label{fig:ad}
\end{figure}

In an adversarial environment, the accuracy of classification has little significance if the deployed classifier can be easily evaded by an intelligent adversary \cite{kantchelian2013approaches}. Classifiers operating in such environments are susceptible to exploratory attacks by an adversary \cite{abramson2015toward}, who uses the same channel as the input data to probe the system inorder to gain information about it. As seen in Fig.~\ref{fig:ad}, a model trained and deployed by a data miner (the \textit{defender}) can provide services to end users, but it is also vulnerable to attacks, which use carefully crafted input samples to evade the classification. In doing so, the classifier system can be viewed as a black box ($C$), providing tacit \textit{Accept/Reject} feedback \cite{papernot2016transferability}. This feedback can be harnessed by an adversary, who is also equipped with the knowledge of machine learning, to reverse engineer ($C'$) the behavior of the black box and avoid detection on future samples. The symmetry of the attack-defense cycle and the new gamut of vulnerabilities introduced by using classification at the core of cybersecurity systems, warrants a data driven analysis of the problem and its effects. 

The security of machine learning has garnered recent interest in literature \cite{abramson2015toward,alabdulmohsin2014adding,smutz2016tree,papernot2016limitations,papernot2016transferability,tramer2016stealing}  . A taxonomy of attacks against machine learning systems was proposed in \cite{barreno2006can}, with Causative and Exploratory attacks being the broad classification of attacks, based on the portion of the data mining process they affect. Causative attacks affect the training data and are aimed at misleading the learned model. These attacks can poison the trained model, but can be prevented by careful curation of the training data \cite{li2014improved} and by using data encryption techniques to safeguard the original training data. Exploratory attacks are more commonplace and dangerous, as they affect the testing phase data, which is often unlabeled and difficult to detect \cite{akhtar2011robustness}. Using the same channels as a client user these attacks can masquerade as regular data samples, posing a risk to any deployed machine learning model. Mimicry attacks\cite{smutz2016tree}, spoofing\cite{akhtar2011robustness} and reverse engineering \cite{alabdulmohsin2014adding} are all forms of exploratory attacks. Exploratory attacks cause the training and testing data distributions to drift, leading to non-stationarity and subsequent degradation in the predictive power of a classifier \cite{kantchelian2013approaches}. These attacks can affect any deployed system, even if the system is available only as a black box service. Recent advancements in black box \textit{Machine-Learning-as a-Service} providers (such as Amazon AWS\footnote{\url{https://aws.amazon.com/machine-learning/}}, Google Cloud Platform\footnote{\url{https://cloud.google.com/prediction/}} and BigML\footnote{\url{https://bigml.com/}}), promise a new era of flexibility and ubiquity in the usage of machine learning. However, preliminary analysis in \cite{tramer2016stealing} has shown that these services are vulnerable to exploratory attacks, by means of querying the system through their APIs. Membership inference attacks presented in \cite{shokri2016membership}, shows that these services are also vulnerable to data leakage, i.e. inferring whether a particular sample belongs to a model's training data, leading to privacy concerns in using machine learning models. The scope of these attacks can be made independent of the model trained and deployed as the black box. Deep neural networks were shown to be vulnerable in \cite{papernot2016limitations}, who later extended their work in \cite{papernot2016transferability} to show how various classifiers, treated as transferable black boxes, are all equally vulnerable to such attacks.

This paper analyzes the vulnerability of classification systems to exploratory attacks, from a data driven perspective. The effects of an adversary, capable of accessing the system only as a black box; without any information about the learning process of the defender's classifier, is presented. To the best of our knowledge, this is the first work which aims at understanding attacks as a exploration-exploitation problem and presents data generation attack algorithms, using the classifier as a black box oracle. The following are the contributions of the proposed work:

\begin{itemize}
\item The vulnerabilities of classifiers operating in adversarial environments, to  probing based attacks, is demonstrated. These attacks show that classification systems should not be naively used in cybersecurity applications, as they can be easily evaded. 
\item A domain independent and data-driven framework is presented, which can be used to simulate attacks on classifiers. Under this general framework,  two specific attack algorithms are provided: the Anchor Points (AP) attacks and the Reverse Engineering (RE) attacks. 
\item Experimental analysis on 10 real world datasets demonstrate that only information about the feature space is sufficient to launch an attack against classifiers, while being agnostic of the type of classifier, the training dataset and the domain of application. This analysis serves as a background for developing  secure machine learning frameworks. 
\end{itemize} 

The rest of the paper is organized as follows. Section~\ref{sec:relatedwork} presents related work in the area of exploratory attacks on classifiers. Data driven attacks on binary classifiers are presented in Section~\ref{sec:datadriven}. Two specific strategies and algorithms are presented in Section 3.1 and 3.2, for generating simple probing attacks to complex reverse engineering attacks. Section~\ref{sec:experiment} presents experimental evaluation on 10 real world datasets, 7 from classification domains and 3 from cybersecurity domains. Avenues for further development are presented in Section~\ref{sec:conclusion}. 

\section{Related Work on Exploratory Attacks on Machine Learning based Classifiers}
\label{sec:relatedwork}

Once a model is trained and deployed in a cybersecurity application, it is vulnerable to exploratory attacks. These attacks are non-intrusive and aim at gaining information about the system, which is then exploited to create evasive samples, to avoid detection. These attacks are universal and are difficult to eliminate by traditional encryption/security
techniques, because they use the same access channels as regular client users and see the same black box view of the system. Work in \cite{nelson2010near}, shows that linear and convex inducing classifier are all vulnerable to probing based exploratory attacks. Exploratory attacks are classified as either: \textit{Targeted} or \textit{Indiscriminate}, based on the specificity of the attacks \cite{barreno2006can}. Targeted attacks aim at modifying a specific set of malicious input samples, minimally, to disguise them as legitimate. Indiscriminate attacks are more general in their goals, as they aim to produce any sample which will avoid detection by the defender's model. Most work on exploratory attacks are concentrated on the targeted case, considering it as
a constrained form of indiscriminate attacks, with the goal of starting with a malicious sample and making minimal modifications to them to avoid detection \cite{biggio2013evasion, lowd2005adversarial}.

Particular strategies developed for performing exploratory attacks vary based on the amount of information available to the adversary, with a broad classification presented in \cite{alabdulmohsin2014adding} as: a) Evasion attacks and b) Reverse Engineering attacks. Evasion attacks are used when limited information about the system is available, such as a few legitimate samples only. An example of evasion is seen in case of the 'Good Words' attacks on spam classifiers, where the word \textit{SALE} is modified to \textit{SA1E}, to avoid being flagged \cite{lowd2005good}. In \cite{xu2016automatically}, a general purpose domain independent evasion technique was developed. Genetic programming was used to generate variants of a set of malicious samples, as per a monotonically increasing fitness function describing success of evasion. This technique is attractive due to its generality, but its practically is limited by the lack of a graded fitness function and limited probing budgets. The gradient descent attacks of \cite{biggio2013evasion} provides an efficient heuristic approach to utilizing the information about the classifier model, to generate optimal modification for targeted evasion of a set of data samples. However, the attack strategy relies on knowing the exact model of the defender and cannot be effectively used when only a black box interface to the defender's classifier is presented. 

Reverse engineering attacks on classifiers, provides avenues for large scale evasion, as it conveys important internal information about the importance of features to the classification task. Reverse engineering was employed in \cite{lowd2005adversarial}, where a signed witness test was used to identify if a particular feature has a positive or a negative impact on the prediction decision. Reverse engineering of a decision tree model was presented in \cite{xu2014evasion}. In \cite{xu2016automatically}, genetic programming was used as a model independent reverse engineering tool, assuming that the training data is known. The idea of reverse engineering was linked to that of active learning in \cite{alabdulmohsin2014adding}.Here, the robustness of Support Vector Machines (SVM) classifiers, to reverse engineering, was tested using active learning techniques of random sampling, uncertainty sampling and selective sampling. 

In the above described targeted-exploratory attacks, it is assumed that an adversary would give up if an attack is expensive (far from the original samples). These attacks do not consider the case of a determined adversary intending to launch an indiscriminate attack. These type of attacks have been largely ignored, with the only mention we found was in \cite{zhou2012adversarial}, where it is termed - the free range attack, as an adversary is free to move about in the data space. Analyzing performance of models under such attack scenarios is essential to understanding its vulnerabilities in a more general and real world situation, where all types of attacks are possible. Also, while most recent methodologies develop attacks as an experimental tool to test their safety mechanisms, there is very few works \cite{papernot2016transferability, shokri2016membership,tramer2016stealing }, which have attempted to study the attack generation process itself. Our proposed work analyzes the vulnerability of classifiers to \textit{Indiscriminate-Exploratory} attacks, where only a black box model of the defender is available. A data driven framework is proposed, thereby highlighting the symmetry of the problem of attack and defense, to motivate a data driven solution.

\section{Data Driven Attacks on Classifiers}
\label{sec:datadriven}

Data driven attacks on classification systems are exploratory in nature. An adversary proceeds by making probes to the classifier, by means of generated input samples, which it presents to the system. The feedback, a simple accept/reject in most cases, can then be used to infer the nature of the trained model. The classifier is seen only as a black box , which can provide binary feedback on input samples,  in the same way as it provides classification on regular benign input data (Fig.~\ref{fig:ad}). As an example, a spam classification system will not provide any information other than its intended behavior of marking input emails as spam, based on its analysis.
In this setting, the model of an adversary can be formalized as follows:
\begin{itemize}
\item \textbf{Knowledge}: Adversary is aware of the number, range and type of features used by the system. This can be approximated from publicly available case studies or by educated guessing. No information about classifier type and training data is known. 
\item \textbf{Goals}: Adversary intends to produce false negatives for the classification. 
\item \textbf{Resources}: The attacker has access to the system as a client user. It can submit probe samples and receive feedback, up to a budget B, without being detected. 
\end{itemize}

Based on the above model of an adversary, the problem of generating exploratory attacks can be formalized. A binary classifier $C$, trained on a set of training data $D$, is deployed to classify input samples as \textit{Legitimate} or \textit{Malicious}. An adversary aims to generate samples $D'_{Attack}$, such that $C(D'_{Attack})$ has a high false negative rate. The adversary has at its disposal a budget $B_{Explore}$ of probing data $D'_{Explore}$, which it can use to learn about $C$ and understand it as $C'$. This setting represents a natural scenario where attackers start with limited reconnaissance and then launch a dedicated campaign, based on the learned vulnerabilities. An adversary operating in this scenario can utilize an \textbf{Explore-Exploit} strategy, popular in search based optimization techniques \cite{wang2015active}, to best utilize the budget $B_{Explore}$ to produce $D'_{Explore}$. Two specific instantiations of this general idea are presented here as the Anchor Points (AP) attack and the Reverse Engineering (RE) attack. 

\subsection{The Anchor Points(AP) Attack}
\label{sec:ap}

Anchor Points attacks start with an adversary gaining information about a set of samples classified as \textit{Legitimate} by the classifier \textit{C}, which it then uses as \textit{Anchors} to launch new attack instances. These attacks are characteristic of an adversary who has a limited probing budget $B_{Explore}$ and who wishes to quickly exploit a new found vulnerability, as is common in the case of zero day exploits\cite{bilge2012before}. 

\begin{figure}[t]
\centering
\includegraphics[width=0.9\textwidth]{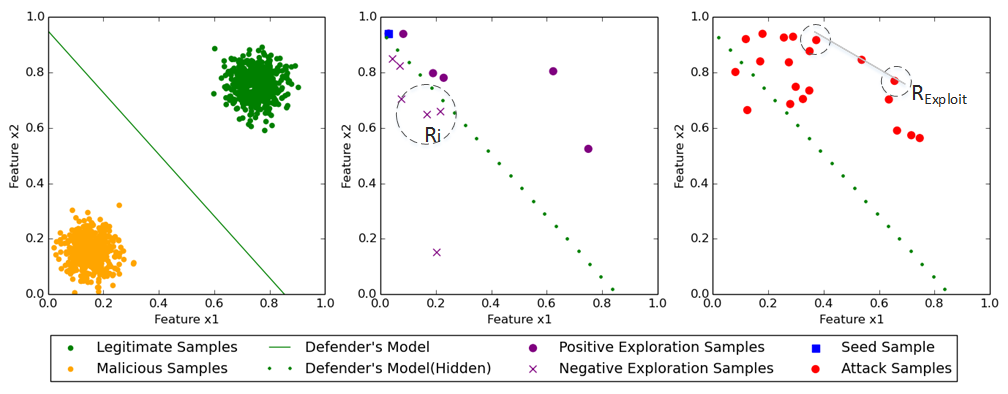}
\caption{Illustration of AP attacks on 2D synthetic data. \textit{(Left - Right)}: The defender's model from training data. The Exploration phase depicting the seed (blue) and the anchor points samples. The Exploitation phase samples generated based on the anchor points, and submitted as attack payload.}
\label{fig:ap}
\end{figure}

From a data driven perspective, the attacks begin by obtaining a set of seed \textit{Legitimate} samples $D'_{Seed}$. As an example, the adversary could start with a set of legitimate emails obtained from its' own inbox, for a spam evasion task. The obtained seed samples are then used to trigger the exploration phase as described in Algorithm~\ref{algo:ap_explore}. The exploration phase intends to obtain a set of \textit{Anchor Points}, which will serve as ground truth for representing the space of samples classified as \textit{Legitimate}. This exploration is performed using a radius based incremental neighborhood search around the seed samples, guided by the feedback from the black box classifier $C$. Diversity of search is ensured by dynamically updating the neighborhood radius $R_i$ in every iteration, as given by (Line 5). This equation causes the radius of exploration to increase in cases where the number of legitimate samples obtained are high, thereby balancing diversity of search with its accuracy. Samples are explored by searching for a random sample within the radius of exploration, as given by (Line 6). The final exploration dataset of \textit{Anchor Points} - $D'_{Explore}$, is comprised of all the explored samples $x_i$ such that $C(x_i)$ is $Legitimate$. This is depicted in Fig.~\ref{fig:ap},  as the set of positive points which are obtained at the end of the exploration cycle. 

\begin{algorithm}[t]
\SetKwInOut{Input}{Input}
\SetKwInOut{Output}{Output}
 \Input{Seed Data $D'_{Seed}$, Defender black box $C$. \textit{Parameters}: Exploration budget $B_{Explore}$,  Exploration neighborhood- [$R_{min}$, $R_{max}$]  }
 \Output{Exploration data set $D'_{Explore}$}
   \setcounter{AlgoLine}{0}

 $D'_{Explore} \leftarrow D'_{Seed}$
 
 count\_ legitimate=0

 \For{ i = 1 .. $B_{Explore}$}{
 
 	$x_i$ $\leftarrow$ Select random sample from $D'_{Explore}$
 	
 	$R_{i} = (R_{max}-R_{min}) * (count\_legitimate / i) + R_{min}$\\
 	
 	$\hat{x_i}\leftarrow$ \textit{Perturb}($x_i$ , $R_{i}$) \Comment{Perturbed sample}\\

	\If{C.predict($\hat{x_i}$) is \textit{Legitimate}}
	{
		$D'_{Explore}\cup$ $\hat{x_i}$
		
		count\_legitimate ++
	}
	}   
\textbf{Procedure} Perturb(sample, $R_{Neigh}$)

$\quad$return sample+=random(mean=0, std=$R_{Neigh}$)
\caption {AP- Exploration Phase}
\label{algo:ap_explore}
\end{algorithm}

\begin{algorithm}[t]
\SetKwInOut{Input}{Input}
\SetKwInOut{Output}{Output}
 \Input{Exploration data set $D'_{Explore}$, Number of attacks $N_{Attack}$, Radius of Exploitation $R_{Exploit}$  }
 \Output{Attacks set $D'_{Attack}$}

 $D'_{Attack}\leftarrow$[]

 \For{ i = 1 .. $N_{Attack}$}{
 
 		$x_A,x_B \leftarrow$ Select random samples from $D'_{Explore}$
 		 				
 		$\hat{x_A},\hat{x_B}\leftarrow Perturb(x_A,R_{Exploit})$, $Perturb(x_B,R_{Exploit})$
 		
		$\lambda=random(0,1)$ \Comment{Random number in [0,1]}

		$attack\_sample_i \leftarrow \hat{x_A}* \lambda + (1-\lambda)*\hat{x_B}$	 
					
		$D'_{Attack}\cup attack\_sample_i$
	}

\textbf{Procedure} Perturb(sample, $R_{Exploit}$)

$\quad$return sample+=random(mean=0, std=$R_{Exploit}$)
\caption {AP- Exploitation Phase}
\label{algo:ap_exploit}
\end{algorithm}

The explored anchor points set $D'_{Explore}$, serves as the basis to launch a dedicated attack campaign. Algorithm~\ref{algo:ap_exploit} incorporates information learned in the exploration phase, to generate the attack samples while also imparting diversity to the attack set $D'_{Attack}$. Diversity is imparted by adding random perturbation to the samples (Line 4)  and then generating a sample based on their convex combination (Line 6), as inspired by the Synthetic Minority Oversampling Technique(SMOTE) for imbalanced datasets \cite{chawla2002smote}. Random perturbation is controlled by the input parameter $R_{Exploit}$, which is kept close to $R_{min}$ (Algorithm~\ref{algo:ap_explore}), as no explicit feedback from the black box $C$, is available in this phase. The final set of attack samples $D'_{Attack}$ is submitted as attack on the classifier $C$, shown as red samples in Fig.~\ref{fig:ap}. An adversary aims to cause a high false negative rate for $C$, while at the same time have high diversity in its attacks. $D'_{Attack}$ is kept much larger than $B_{Explore}$ to justify adversarial budget expenditure. 

The performance of the Anchor Points attack depends on the diversity and accuracy of samples collected in the exploration phase. Larger coverage ensures flexibility in the attack phase. By the nature of these attacks, they could be thwarted by blacklists capable of approximate matching \cite{prakash2010phishnet}. Nevertheless, these techniques are suited for adhoc swift exploits, to cause impact before a defender has time to respond.

\subsection{The Reverse Engineering(RE) Attack}
\label{sec:re}

These attacks aim to directly reverse engineer the classification boundary, so as to better understand the classification landscape, which can then be leveraged to launch large scale evasion attacks on the black box $C$. Reverse engineering could be a goal in itself, as it provides information about features importance to the classification task, or it could be a first step to launching an evasion or availability attack \cite{alabdulmohsin2014adding}. A reverse engineering attack, if done effectively, can avoid detection and can make retraining more difficult on the part of the defender. However, unlike the AP attack, the reverse engineering process is affected by the type of classifier model used by $C$, and is also dependent on the availability of sufficient exploration budget $B_{Explore}$, for the reverse engineering learning task. Nevertheless, an adversary motivated to evade the classification system is not concerned with fitting the decision boundary of $C$ exactly. A linear approximation to the non linear defender's boundary is sufficient to launch a reduced accuracy attack,  which can be compensated for by launching a massive attack campaign, utilizing the information provided by the reverse engineered model.

\begin{figure}[t]
\centering
\includegraphics[width=0.9\textwidth]{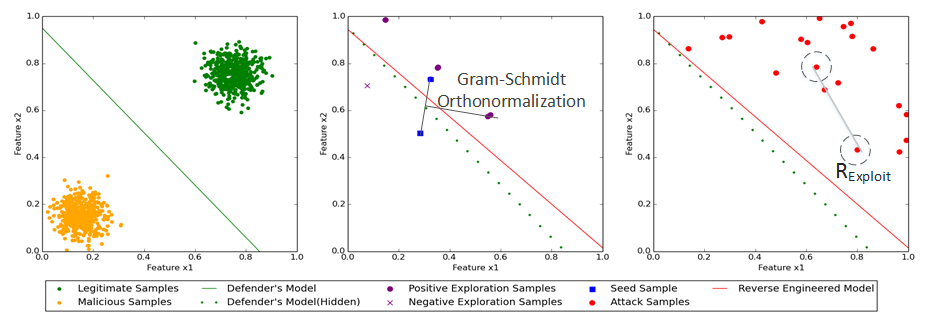}
\caption{Illustration of RE attacks on 2D synthetic data.\textit{(Left - Right)}: The defender's model based on training data. The Exploration phase depicting reverse engineering using the Gram-Schmidt orthonormalization process. The Exploitation attack phase samples generated after validation from the surrogate classifier.}
\label{fig:re}
\end{figure}

Effective reverse engineering depends on making best use of the $B_{Explore}$. Random sampling can lead to wasted probes, with no new information added. The query synthesis strategy of \cite{wang2015active} generates samples close to the decision boundary and spreads these samples across the boundary, for better learning of the decision landscape. The approach in \cite{wang2015active} was used for selecting samples for active learning. We modify the approach for the task of reverse engineering in Algorithm~\ref{algo:re_explore}, where a surrogate classifier $C'$ is learned as a result of the exploration phase. The algorithm begins by accepting a seed datasets, which is comprised of atleast one \textit{Legitimate} and \textit{Malicious} sample. The algorithm then employs the Gram-Schmidt process \cite{wang2015active}, to generate orthonormal samples near the midpoint of two randomly selected points of the opposite classes, as shown in Fig.~\ref{fig:re}. The magnitude of the orthonormal mid-perpendicular vector is set to $\lambda_i$, selected as random value in [0,$\lambda$], to incorporate variability in the exploration phase (Line 8). The resulting exploration samples are then classified as Legitimate/Malicious by $C$, which can be probed upto $B_{Explore}$. The final combined dataset $D'_{Explore}$ (Line 14) is then used to train a linear classifier of choice, to form the surrogate reverse engineered model $C'$ (Line 15).

\begin{algorithm}[t]
\SetKwInOut{Input}{Input}
\SetKwInOut{Output}{Output}
 \Input{Seed Data $D'_{Seed}$, Defender black box model $C$. \textit{Parameters}: Exploration budget $B_{Explore}$, Magnitude of dispersion $\lambda$}
 \Output{Exploration data Set $D'_{Explore}$, Surrogate classifier $C'$}

 $D'_{Explore\_L}$, $D'_{Explore\_M}$ = Legitimate, Malicious samples of $D'_{Seed}$

 \For{ i = 1 .. $B_{Explore}$}{
 		$x_L$, $x_M$ $\leftarrow$ Select random samples from  $D'_{Explore\_L}$, $D'_{Explore\_M}$
 	 	
 	 	$x_0$=	$x_L-x_M$
 	 	
 	 	Generate random vector $x_R$
 	 	
		$x_R=x_R-\frac{<x_R,x_0>}{<x_0, x_0>}*x_0$  \Comment{Gram-Schmidt process - $x_R$ orthogonal to $x_0$}
		
		$\lambda_i=random(0,\lambda)$		
		
		$x_R$=$\frac{\lambda_i}{norm(x_R)}$*$x_R$ \Comment{Set magnitude of orthogonal midperpendicular}
		
		$x_S$=$x_R+(x_L+x_M)/2$ \Comment{Set $x_R$ to midpoint}
		
		\If{C.predict($x_S$) is Legitimate}{
		$D'_{Explore\_L} \quad \cup \quad x_S$
		}
		\Else{
				$D'_{Explore\_M} \quad \cup \quad x_S$
		}
	}
	$D'_{Explore}$ =	$D'_{Explore\_L} \quad \cup\quad D'_{Explore\_M}$
	
	Train $C'$ using $D'_{Explore}$ \Comment{Training can be based on linear classifier of choice}   

\caption {RE Exploration - Using Gram-Schmidt process.}
\label{algo:re_explore}
\end{algorithm}

The reverse engineered model $C'$ can be used to crosscheck the randomly generated samples in the exploitation phase, to ensure that attacks have high accuracy. A practical and effective exploitation strategy is to use the $D'_{Explore}$ as the seed set for Algorithm~\ref{algo:ap_explore}, with the exception that we use $C'$ to probe instead of the original $C$. Since the $C'$ is a locally trained model, probing it does not impact $B_{Explore}$. Thereby allowing an adversary to make, theoretically, infinite queries to $C'$, at effectively zero cost.  The anchor points obtained can then be used to perform exploitation using Algorithm~\ref{algo:ap_exploit}. The exploitation can use a large $R_{Exploit}$, as the results can be verified against the surrogate $C'$, to ensure higher diversity and higher accuracy of attacks, than the AP attacks. 

\section{Empirical Evaluation}
\label{sec:experiment}

\subsection{Experimental Setup}
\label{sec:setp}

Experimental evaluation, presented here, shows an adversary's point of view of the classification system. The adversary is capable of generating data driven attacks on the system, by making limited probes to it and then generating a dedicated campaign of evasive samples. Since the adversary aims at evading the classification system, its efficacy is measured as the Effective Attack Rate(EAR), given by Equation~\ref{eqn:EAR}. This equation measures the accuracy of the attacks, from an adversary's point of view, and it indicates the false negative rate of the defender's classifier on the attack samples. 

\begin{equation}
EAR\quad =\quad \frac { \left| \left\{ x:\quad C(x) = Legitimate\quad \wedge \quad x\in D'_{Attack} \right\} \right|  } {\left| D'_{Attack} \right|}
\label{eqn:EAR}
 \end{equation}

Here, $C$ represents the defender's black box classifier and $D'_{Attack}$ is the set of attack campaign samples generated by the adversary to attack $C$. The goal of an adversary is to maximize EAR. 

The evaluation is performed using 10 binary classification datasets, shown in Table~\ref{tbl:linear} (\textit{Column 1}). The first 7 datasets represent standard classification tasks and were obtained from the UCI machine learning repository \cite{Lichman:2013}. The Spambase dataset for email classification \cite{Lichman:2013}, the KDD99 intrusion detection dataset\cite{Lichman:2013} and the CAPTCHA dataset for classifying human-bots based on behavioral data \cite{d2014avatar}, represent 3 different cybersecurity applications which employ machine learning based classification at its core. All datasets were transformed to have numerical values normalized in the range of [0,1]. Instances were shuffled to remove any bias due to inherent concept drift. The class label 1 is taken as the \textit{Malicious} class and 0 is taken as the \textit{Legitimate} class, as convention.

In all experiments, the exploration probe budget $B_{Explore}$ is taken as 1000 and the number of attack samples to be generated $N_{Attack}$ is taken as 2000. For the Anchor Points (AP) attack, the neighborhood radius [$R_{min}, R_{max}$] is taken as [0.1,0.5] and the exploitation radius as $R_{Exploit}$ = 0.1. In case of the reverse engineering (RE) attacks, a larger exploitation radius ($R_{Exploit}$=0.5) is considered, due to additional validation provided by the surrogate learned classifier $C'$. A SVM with linear kernel and high regularization constant (c=10) is taken for the surrogate classifier,  to prevent overfitting to $D'_{Explore}$. The magnitude of dispersion ($\lambda$) is taken as 0.25, and it was found that changing this had little affect on the final results. In all experiments, no information about the black box $C$ is known by the adversary. All experiments in this section are performed using Python 2.7 and the scikit-learn library\cite{pedregosa2011scikit} . Results are averaged over 30 runs for every experiment. Section~\ref{sec:linear} presents the results of a symmetric case, where both the adversary and the defender have similar model types, while Section~\ref{sec:nonlinear} presents analysis on non symmetric model types, with 4 different classifiers for the black box.

\begin{table}[t]
\centering
\caption{Results of AP and RE attacks on a linear defender model on 10 real world datasets (EAR - Effective Attack Rate). }
\label{tbl:linear}
\begin{tabular}{|l|c|c|c|c|c|}
\hline
\multicolumn{1}{|c|}{\multirow{3}{*}{\begin{tabular}[c]{@{}c@{}}Dataset \\ (\#Instances, \\ \#Attributes)\end{tabular}}} & \multirow{3}{*}{\begin{tabular}[c]{@{}c@{}}Defender's\\ Initial \\ Accuracy\end{tabular}} & \multirow{3}{*}{\begin{tabular}[c]{@{}c@{}}Explored\\  Anchor\\  Points/$B_{Explore}$\end{tabular}} & \multirow{3}{*}{\begin{tabular}[c]{@{}c@{}}Accuracy of\\  RE model C'\end{tabular}} & \multicolumn{2}{c|}{EAR} \\ \cline{5-6} 
\multicolumn{1}{|c|}{} &  &  &  & \multirow{2}{*}{AP} & \multirow{2}{*}{RE} \\
\multicolumn{1}{|c|}{} &  &  &  &  &  \\ \hline
Digits08 (1500,16) & 0.98 & 0.63 & 0.92 & 0.96$\pm$0.01 & 0.93$\pm$0.06 \\ \hline
Credit (1000,61) & 0.79 & 0.71 & 0.71 & 0.98$\pm$0.01 & 0.8$\pm$0.15 \\ \hline
Cancer (699, 10) & 0.97 & 0.99 & 0.95 & 0.99$\pm$0.01 & 0.99$\pm$0.01 \\ \hline
Qsar (1055, 41) & 0.87 & 0.99 & 0.42 & 0.99$\pm$0.01 & 0.99$\pm$0.01 \\ \hline
Sonar (208, 60) & 0.88 & 0.98 & 0.61 & 0.99$\pm$0.01 & 0.98$\pm$0.01 \\ \hline
Theorem (3060, 51) & 0.72 & 0.67 & 0.57 & 0.97$\pm$0.01 & 0.87$\pm$0.08 \\ \hline
Diabetes (768, 8) & 0.78 & 0.5 & 0.71 & 0.98$\pm$0.01 & 0.95$\pm$0.04 \\ \hline
Spambase (4600, 57) & 0.91 & 0.5 & 0.59 & 0.93$\pm$0.01 & 0.71$\pm$0.2 \\ \hline
KDD99 (494021, 41) & 0.99 & 0.91 & 0.55 & 0.99$\pm$0.01 & 0.93$\pm$0.04 \\ \hline
CAPTCHA (1885, 26) & 1.0 & 0.92 & 0.91 & 0.99$\pm$0.01 & 0.97$\pm$0.02 \\ \hline
\end{tabular}
\end{table}

\subsection{Experiments with Symmetric Defender Model}
\label{sec:linear}

Experiments with a linear kernel SVM for the defender's model (regularization parameter, c=1), is presented here, to show effects of a symmetric model type between the adversary and the defender. The initial accuracy of the defender, as perceived by cross-validation on its training dataset before deployment, is shown in Table~\ref{tbl:linear}. The Effective Attack Rate (EAR) shows that even models which are perceived to have a high accuracy ($>$70\% in all 10 cases) by the defender, are effectively evaded by an adversary, with an EAR of 97.7\% in case of the AP attacks and 91.2\% for the RE attacks, on average. This shows the inherent vulnerability of the classification models and the misleading nature of accuracy, in an adversarial environment. 

Both the attack methods of AP and RE are effective. The fundamental difference between the two approaches is that RE places its confidence in its understanding of the separating boundary, while the AP approach places its confidence only on the anchor points obtained during exploration. From Table~\ref{tbl:linear}, it is seen that the number of anchor points obtained is $>$50\% of $B_{Explore}$(\textit{Column 3}), making it a simple attack strategy in high dimensional spaces. For the RE attack, the accuracy of the surrogate classifier $C'$(\textit{Column 4}) is computed by evaluating it on the original dataset, as an adhoc metric of $C'$s understanding of the original data space and the extent of reverse engineering. It is observed that even in cases where the RE accuracy is low (0.41 for Qsar), a high EAR is seen (0.99). This is because, the goal of the adversary is not to totally reverse engineer the model C, but instead to learn it sufficiently enough to evade it. This enables the linear approximation technique of the RE approach to work as an effective attack strategy. The higher variability in EAR for the RE attacks is also a result of this approximation and its dependence on the quality of the exploration data. 

\begin{figure}[t]
\centering
\subfloat[Effect of $R_{Exploit}$ on EAR]{\includegraphics[width=0.4\linewidth, height=0.22\textwidth]{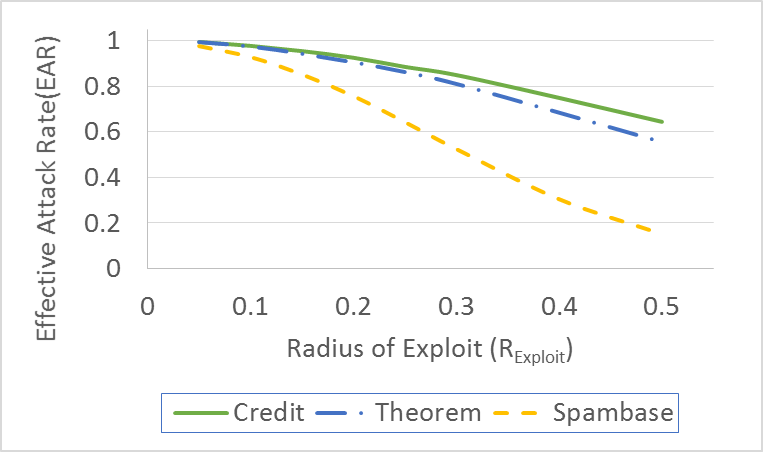}}\quad 
\subfloat[Effect of $B_{Explore}$ on EAR]{\includegraphics[width=0.4\linewidth, height=0.22\textwidth]{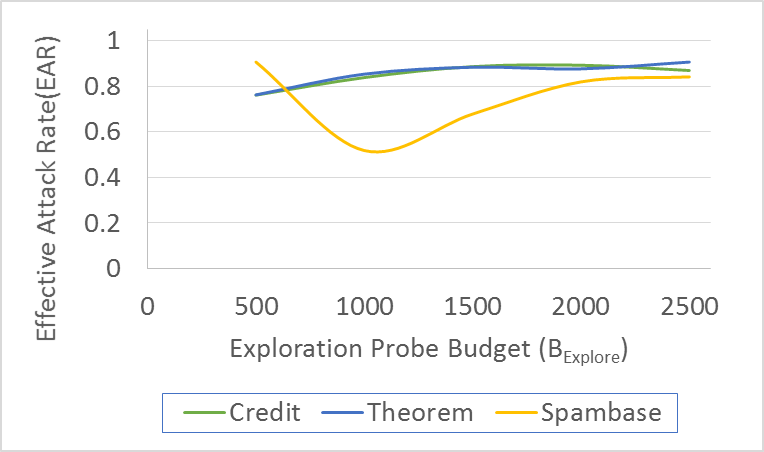}}
\caption{Improving diversity of attacks in AP leads to decreased EAR (a). In case of RE, which has high diversity, EAR can be improved by increasing $B_{Explore}$(b).  }
\label{fig:explore_budget}
\end{figure}

While the AP attacks have a higher EAR than the RE attacks (Table~\ref{tbl:linear}), the RE approach provides better diversity of attacks, as it uses a larger $R_{Exploit}=0.5$ (for AP,$R_{Exploit}=0.1$ ). Diversity ensures that simple countermeasures of blacklisting will not thwart an attack \cite{prakash2010phishnet}. The effect of increasing the $R_{Exploit}$ for the AP attack, in an attempt to increase its diversity, is shown in  Fig.~\ref{fig:explore_budget}a). It is seen that this leads to rapid deterioration in the EAR, as the only ground truth information available is the Anchor Points obtained during exploration. Increasing distance from these points leads to uncertainty and reduced accuracy of attacks. For the RE attacks, the datasets of - Cancer, Theorem and Spambase, are seen to have a low EAR in Table~\ref{tbl:linear}. Effects of  increasing the $B_{Explore}$,  in an attempt to increase the EAR is shown in Fig.~\ref{fig:explore_budget}b). Increasing number of probes leads to an increase in the understanding of the black box $C$, which translates to better EAR. This increase plateaus after a critical mass of samples, needed for active learning the model $C$, is reached. This indicates the efficacy of the RE attacks as a long term attack strategy on a classification system, where over time the additional information learned can lead to both high accuracy and high diversity. 

\subsection{Experiments with Non-Symmetric Defender Model}
\label{sec:nonlinear}

The development of the AP and the RE attack strategies considers a black box defender's model. These techniques are essentially data space search approaches, which are independent of the defender's underlying model type and their parameters. Results of testing the behavior of these attack strategies on a non-symmetric and non-linear defender models is presented in Table~\ref{tbl:non_linear}. The following defender black box models are considered: K-Nearest Neighbor classifier (KNN) with K=3, SVM with a radial basis function kernel(SVM-RBF, $\gamma=$0.1), C4.5 decision trees (DT), and random forest of 50 decision models (RF) \cite{pedregosa2011scikit}. The parameters of the experiments are kept the same as in Section~\ref{sec:linear}.

\begin{table}[t]
\centering
\caption{Effective Attack Rate (EAR) of AP and RE attacks, with non linear defender's model (Low EAR values are italicized.)}
\label{tbl:non_linear}
\begin{tabular}{l|c|c|c|c|c|c|c|c|}
\cline{2-9}
\multicolumn{1}{c|}{\textbf{}} & \multicolumn{2}{c|}{\textit{\textbf{KNN}}} & \multicolumn{2}{c|}{\textit{\textbf{SVM-RBF}}} & \multicolumn{2}{c|}{\textit{\textbf{DT}}} & \multicolumn{2}{c|}{\textit{\textbf{RF}}} \\ \hline
\multicolumn{1}{|l|}{\textit{\textbf{Dataset}}} & \textit{\textbf{AP}} & \textit{\textbf{RE}} & \textit{\textbf{AP}} & \textit{\textbf{RE}} & \textit{\textbf{AP}} & \textit{\textbf{RE}} & \textit{\textbf{AP}} & \textit{\textbf{RE}} \\ \hline
\multicolumn{1}{|l|}{Digits08} & 0.89 & 0.96 & 0.97 & 0.89 & 0.87 & 0.63 & 0.85 & \textit{0.48} \\ \hline
\multicolumn{1}{|l|}{Credit} & 0.96 & 0.78 & 0.94 & 0.53 & 0.79 & \textit{0.42} & 0.79 & \textit{0.33} \\ \hline
\multicolumn{1}{|l|}{Cancer} & 0.99 & 0.99 & 0.99 & 0.99 & 0.97 & 0.89 & 0.99 & 0.98 \\ \hline
\multicolumn{1}{|l|}{Qsar} & 1 & 0.99 & 0.99 & 0.99 & 0.96 & 0.76 & 0.99 & 0.99 \\ \hline
\multicolumn{1}{|l|}{Sonar} & 0.99 & 0.98 & 1 & 1 & 0.97 & 0.62 & 0.99 & 0.95 \\ \hline
\multicolumn{1}{|l|}{Theorem} & 0.97 & 0.813 & 0.95 & 0.5 & 0.95 & 0.79 & 0.62 & 0.78 \\ \hline
\multicolumn{1}{|l|}{Diabetes} & 0.99 & 0.935 & 0.99 & 0.9 & 0.83 & 0.63 & 0.88 & 0.61 \\ \hline
\multicolumn{1}{|l|}{Spambase} & 0.93 & 0.99 & \textit{0.48} & 0.84 & \textit{0.08} & \textit{0.11} & 0.99 & 0.98 \\ \hline
\multicolumn{1}{|l|}{KDD99} & 0.99 & 0.93 & 1 & 0.99 & 0.89 & 0.54 & 0.92 & \textit{0.27} \\ \hline
\multicolumn{1}{|l|}{CAPTCHA} & 0.99 & 0.92 & 0.99 & 0.92 & 0.97 & 0.83 & 0.93 & 0.89 \\ \hline
\end{tabular}
\end{table}

It is seen that the AP approach is minimally affected by the choice of the defender's model. The RE attacks are affected by the choice of the model, as seen in the case of the Credit and the Theorem datasets. These datasets were seen to have a low accuracy, when trained using a linear model (Table~\ref{tbl:linear}), indicating non linearity in their model space. In these cases, the linear approximation in the RE approach, is not sufficient to have a high EAR. However, in a majority of the cases it is seen that a $>$50\% attack rate is still possible with just a linear SVM model used for reverse engineering. A high average attack rate, irrespective of the underlying classifier used, indicates vulnerability of classification to purely data driven attacks. An interesting observation warranting further investigation is the comparatively low EAR on the decision tree models, which could be indicative of their attack resistance and its inverse relation to model robustness.

\section{Conclusion and Future Work}
\label{sec:conclusion}

We present a general data driven framework for demonstrating the vulnerability of classification systems to exploratory attacks at test time. Under this framework, two specific attack algorithms were developed: The Anchor Points attack (AP) and the Reverse Engineering attacks (RE). The effectiveness of these attack algorithms on 10 real world datasets, demonstrates that adversarial attacks can be launched having only the knowledge of the feature space of the data, agnostic of the defender's classifier type, training data and the domain of application. The defender's perceived accuracy was shown to be of little importance, if the model can be easily evaded by such probing based attacks. This is especially relevant in cybersecurity application domains, where the primary purpose of the classifier is to provide security. In these domains, it is worth emphasizing that - higher accuracy in machine learning does not necessarily imply better security.

While the proposed work presents the adversary's point of view of the attack-defense cycle, its goal is to move towards a more secure paradigm of using classifiers in cybersecurity domains, by clearly understanding its vulnerabilities. Future work will include attack detection and effective relearning, in environments with adversarial activity. 

\bibliographystyle{splncs03}
\bibliography{references}

\end{document}